\documentclass[10pt,twocolumn,letterpaper]{article}

\usepackage{btas}
\usepackage{times}
\usepackage{epsfig}
\usepackage{graphicx}
\usepackage{amsmath}
\usepackage{amssymb}

\usepackage[listofformat=parens, justification=centering, subrefformat=subparens]{subfig}
\usepackage{footmisc} 
\usepackage{tabularx}

\usepackage[pagebackref=true,breaklinks=true,letterpaper=true,colorlinks,bookmarks=false]{hyperref}

\usepackage{fancyhdr}

\btasfinalcopy 

\newcommand\capt[3]{\caption[#2]{\label{#1}\textsc{#2}. \small#3}}
\graphicspath{{./images/}{./figures/}}

\newcommand\ic[2][1]{\includegraphics[width=#1\textwidth]{#2}}
\newcommand\ich[2][1]{\includegraphics[height=#1\textheight]{#2}}
\newcommand\icp[3][1]{\includegraphics[width=#1\textwidth,page=#2]{#3}}
\renewcommand\etal[1]{\textit{et al.}~\cite{#1}}
\renewcommand\sec[1]{Sec.~\ref{sec:#1}}
\newcommand\fig[1]{Fig.~\ref{fig:#1}}
\newcommand\sfig[1]{Fig.~\subref{fig:#1}}
\newcommand\tab[1]{Tab.\;\ref{tab:#1}}

\DeclareMathOperator*{\argmax}{\mathrm{arg\,max}}

\ifbtasfinal\fi
\begin{document}

\title{Unconstrained Face Detection and Open-Set Face Recognition Challenge}

\author{
\normalsize M. G\"unther,$^a$ P. Hu,$^b$ C. Herrmann,$^c$ C. H. Chan,$^d$ M. Jiang,$^e$ S. Yang,$^f$ A. R. Dhamija,$^a$ D. Ramanan,$^b$ \\
\normalsize J. Beyerer,$^c$ J. Kittler,$^d$ M. Al Jazaery,$^e$ M. I. Nouyed,$^e$ G. Guo,$^e$ C. Stankiewicz,$^f$ and T. E. Boult$^a$\\
\small $^a$University of Colorado Colorado Springs, $^b$Carnegie Mellon University Pittsburgh, $^c$Karlsruhe Institute of Technology, \\
\small $^d$University of Surrey, $^e$West Virginia University, $^f$University of Wolverhampton
}

\maketitle

{
  \chead{\footnotesize This is an \textbf{ERRATA} version of the original paper presented at the International Joint Conference on Biometrics (IJCB) 2017.}
  \lhead{}
  \thispagestyle{fancy}
  \pagenumbering{gobble}
}

\begin{abstract}
  Face detection and recognition benchmarks have shifted toward more difficult environments.
  The challenge presented in this paper addresses the next step in the direction of automatic detection and identification of people from outdoor surveillance cameras.
  While face detection has shown remarkable success in images collected from the web, surveillance cameras include more diverse occlusions, poses, weather conditions and image blur.
  Although face verification or closed-set face identification have surpassed human capabilities on some datasets, open-set identification is much more complex as it needs to reject both unknown identities and false accepts from the face detector.
  We show that unconstrained face detection can approach high detection rates albeit with moderate false accept rates.
  By contrast, open-set face recognition is currently weak and requires much more attention.
\end{abstract}

\section{Introduction}
\label{sec:introduction}

Automatic face recognition is an important field and has a tremendous impact on many domains of our life.
For example, private images can be sorted by persons that appear on them (e.g., Apple Photos or Google Photos), or airports perform automatic face recognition as passport control \cite{sanches2016automated}.
As the latter has severe security implications, most face recognition challenges such as the Face Recognition Vendor Tests\footnote{\scriptsize\url{https://www.nist.gov/programs-projects/face-recognition-vendor-test-frvt}} evaluate algorithms that perform verification, i.e., where a pair of model and probe images is tested whether they contain the same identity.
Usually, a similarity between model and probe image is thresholded, where the threshold is computed based on a desired false acceptance rate.
Other challenges included more difficult data, such as the Point and Shoot Challenge \cite{beveridge2015report} or the Face Recognition Evaluation in Mobile Environment \cite{guenther2013face}.

On the other hand, identification seems to be a more intricate problem, as a probe image must be compared to all identities enrolled in a gallery.
As Klontz and Jain \cite{klontz13case} and Kemelmacher-Shlizerman \etal{kemelmacher2016megaface} showed, when the gallery is large and probe images are taken in uncontrolled conditions, identifying the correct person is not trivial.
In real surveillance scenarios subjects usually do not realize that their faces are captured and, hence, do not cooperate with the system.
Furthermore, most of the captured faces will not belong to any person in the gallery and should be declared as unknown, leading to open-set face recognition.
Also, face detectors might have false accepts, i.e., where a region of the background is detected as a face.
These misdetections also need to be classified as unknown by face recognition algorithms.
Therefore, additionally to identifying the correct person in the gallery based on difficult imagery, for an unknown face or misdetection, the similarity to \emph{all} persons in the gallery must be below a certain threshold, which is usually computed based on a desired false identification rate.
While the latest face recognition benchmark IJB-A \cite{klare2015ijba} includes an open-set protocol, it does not treat misdetections that are subsequently labeled with an identity as an error, which makes that benchmark incomplete.

For the UCCS unconstrained face detection and open-set face recognition challenge\footnote{\scriptsize\url{http://vast.uccs.edu/Opensetface}} we invited participants to submit results of face detection and face recognition algorithms.
Given a set of images in the training set, containing 23,349 labeled faces of 1085 known and a number of unknown persons, participants were to detect all faces in the test set, and to assign each detected face an identity of the gallery, or an \emph{unknown} label when the algorithm decided that the person has not been labeled as known.

\section{Dataset}
\label{sec:dataset}
\begin{figure*}[t!]
  \vspace*{-2ex}
  \centering
  \subfloat[Example Images\label{fig:UCCS:examples}]{
    \begin{tabular}{c@{\qquad}c}
      \includegraphics[width=.4\textwidth]{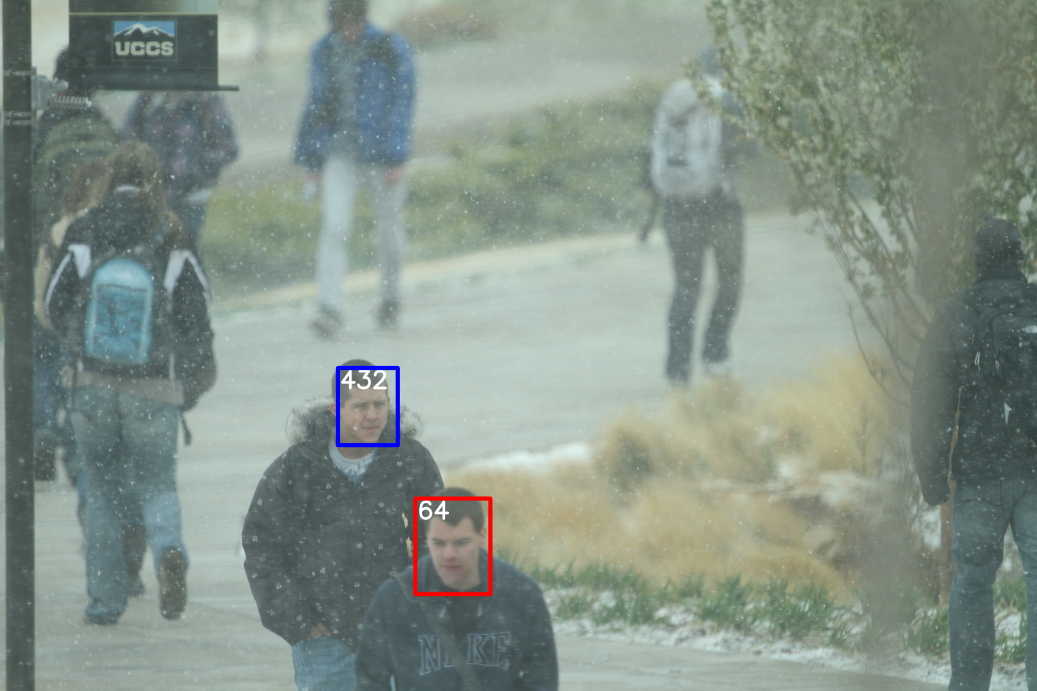} &
      \includegraphics[width=.4\textwidth]{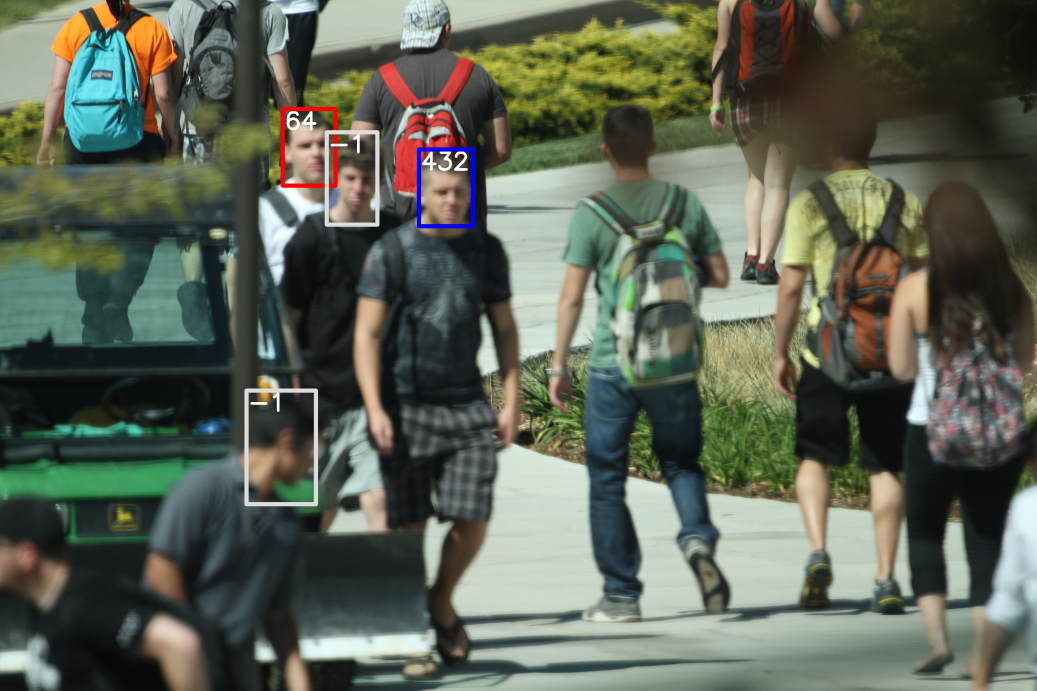}
    \end{tabular}
  }\\
  \subfloat[Example Faces\label{fig:UCCS:faces}]{
    \begin{tabular}{cccccc}
      \includegraphics[height=.12\textheight]{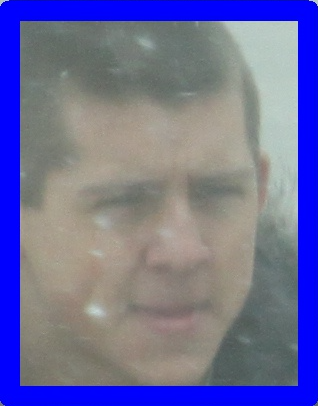} &
      \includegraphics[height=.12\textheight]{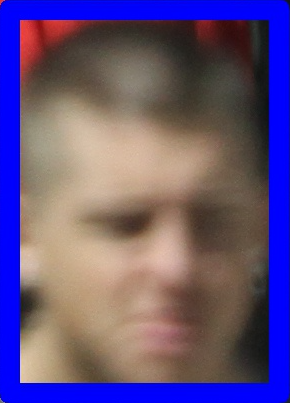} &
      \includegraphics[height=.12\textheight]{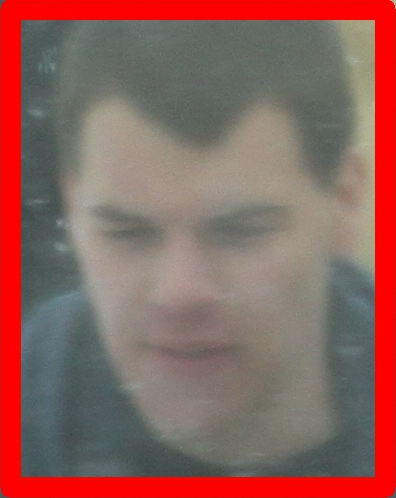} &
      \includegraphics[height=.12\textheight]{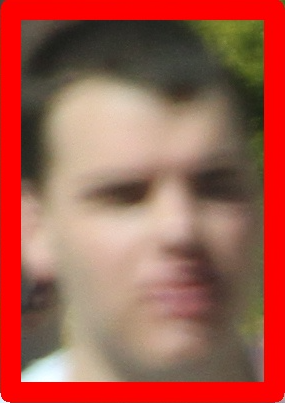} &
      \includegraphics[height=.12\textheight]{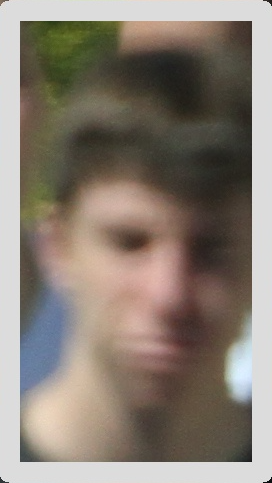} &
      \includegraphics[height=.12\textheight]{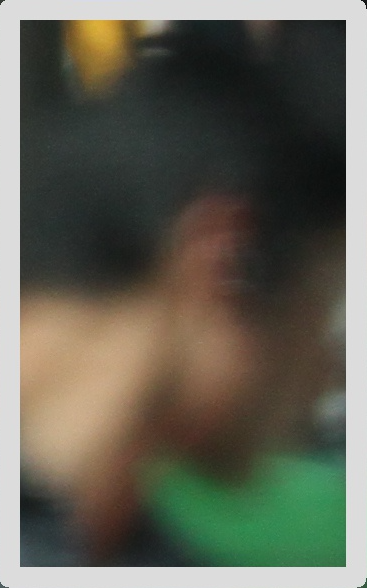}
    \end{tabular}
  }
  \capt{fig:UCCS}{Examples of the UCCS Dataset}{Two exemplary images of the UCCS dataset including hand-annotated bounding boxes and identity labels are shown in \subref*{fig:UCCS:examples}. In \subref*{fig:UCCS:faces} the cropped faces of the two images are displayed. Faces with the same color mark the same identity, while gray boxes mark unknown identities.}
\end{figure*}

To run the challenge, the UnConstrained College Students (UCCS) dataset was developed as a significantly extended version of  the dataset presented by Sapkota and Boult \cite{sapkota2013largescale}.
It contains high-resolution images captured from an 18 megapixel camera at the University of Colorado Colorado Springs, aimed at capturing people walking on a sidewalk from a long range of 100--150 meters, at one frame per second.
The dataset collection was spread across 20 different days, between February 2012 and September 2013 providing a variety of images in different weather conditions such as sunny or snowy days.
There are frequent occlusions due to tree branches or poles as well as sunglasses, winter caps, or fur jackets that make both detection and recognition a challenging problem.
Since the captured subjects are unaware of the dataset collection and casually focus on random activities such as glancing at a mobile phone or conversing with peers while walking, there is a wide variety of face poses along with some cases of motion blur, and many cases where faces are not in the focus of the camera.

The dataset consists of more than 70000 hand-cropped face regions, which are generally larger than the actual face.
An identity is manually assigned to many of the faces, where 20\,\% of these identities appear in two or more days.
Due to the manual nature of the labeling process, for approximately 50\,\% of the face regions no identity could be assigned.
Two example images including their manually cropped and labeled faces are shown in \fig{UCCS}.

\subsection{Protocol}
We split the UCCS database into training, validation and test sets.
For each image in the training and validation sets we provide a list of bounding boxes with their corresponding identity labels, including the label $-1$ for unknown subjects.
In the test set we only supply a list of images, in which the participants need to detect the faces (face detection challenge) and provide an identity label including a similarity score to each bounding box (face recognition challenge).

To be able to evaluate the difference of recognizing unknown identities that have been seen during training (so-called \emph{known unknowns}) and subjects that have never been seen (\emph{unknown unknowns}) \cite{guenther2017toward}, we artificially \emph{mask} several of the known identities.
Some of these masked identities are present in the training, validation and test set, while some other masked identities are excluded from the training set.
More details about the distribution of images and identity labels in our evaluation protocol can be found in \tab{Distribution}.

\newcolumntype{C}{X<\centering}

\begin{table*}[t!]
  \vspace*{-2ex}
	\centering\small
	\begin{tabularx}{.95\textwidth}{|l||c|c||C|C||C|C|}
		\hline
		& \multicolumn{2}{c||}{\textbf{Training}}& \multicolumn{2}{c||}{\textbf{Validation}}& \multicolumn{2}{c|}{\textbf{Test}}\\\cline{2-7}
		& Subjects & Faces & Subjects & Faces & Subjects & Faces\\\hline\hline
		Known & 1085 & 11012 & 990 (1002) & 3004 (3359) & 921 (954) & 12636 (15312)\\\hline
		Unknown & ? & 12156 & ? & 6688 (7575) & ? & 17774 (18983)\\\hline
		Masked in Training & 116 & 181 & 102 & 228 & 116 & 1277\\\hline
		Masked not in Training & 0 & 0 & 461 & 1189 & 526 & 4466\\\hline\hline
		{\bf Total} & 1201 & 23349 & 1553 (1565) & 11109 (12351) & 1563 (1596) & 36153 (40038)\\\hline
	\end{tabularx}
	\capt{tab:Distribution}{Distribution of Subjects and Labeled Faces}{These numbers of subjects and faces are present in the evaluation protocol. The numbers in parentheses display the updated protocol (cf.~\sec{cleanup}). Known subjects were labeled with their (positive) ID, while unknown subjects are labeled as $-1$. For the masked identities, the participants were given the label $-1$, they could not differentiate between them and the unknown identities.}
\end{table*}

\section{Challenge Participants}
\label{participants}

Participants were invited to submit a short description of their algorithms.
They are listed in the order of submission and marked with their according institution ($^a$ -- $^f$, cf.~list of authors on first page).

\def\pphead#1{{\bf #1}:}

\subsection{Face Detection}
\label{sec:participants:detection}

\pphead{Baseline}
The Baseline face detector uses the out-of-the-box face detector of Bob \cite{anjos2012bob}, which relies on boosted Local Binary Pattern (LBP) features \cite{atanasoaei2012multivariate} extracted from gray level images, and is trained on a combination of publicly available close-to-frontal face datasets.
The implementation can be downloaded from the Python Package Index.\footnote{\scriptsize\label{fn:baseline}\url{http://pypi.python.org/pypi/challenge.uccs}}

\pphead{TinyFaces$^b$}
The TinyFaces face detector \cite{hu2017finding} consists of a set of scale-specific mini-detectors, each of which is tuned for a predefined object size and sits on top of a fully convolutional~\cite{long2015fully} ResNet101~\cite{he2015deep}.
Each mini-detector is implemented as a convolutional filter, which takes convolutional features extracted from multiple layers as input and outputs a spatial heat map that represents detection confidence at every location.
In addition, four filters have been tied to each mini-detector for bounding box regression~\cite{ren2015faster}.
The TinyFaces detector is trained on the training set of WIDER FACE~\cite{yang2016wider}.
During training multi-resolution sampling, balanced sampling~\cite{ren2015faster}, and hard negative mining are applied.
During testing the detector works on an image pyramid, while only running mini-detectors tuned for small object size on the interpolated level.
The TinyFaces algorithm was run by the LqfNet$^c$ team.

\pphead{UCCS$^a$}
The UCCS face detector is based on the Baseline face detector from Bob \cite{anjos2012bob}.
Additionally to the LBP features extracted from gray-level images, color information is added in terms of converting the image to HSV color space and extracting quantized hue (H) and saturation (S) values.
A face detection cascade is trained on a combination of LBP and color values, using the training set images of the MOBIO \cite{mccool2012mobio}, SCface \cite{grgic2011scface}, and CelebA \cite{liu2015deep} datasets, as well as the training images of the UCCS dataset.

\pphead{MTCNN$^d$}
Faces and facial landmarks are detected by MTCNN~\cite{zhang2016joint}.
Besides increasing the minimally detectable face bounding box to 75 pixels, the publicly available face detector\footnote{\scriptsize\url{http://github.com/kpzhang93/MTCNN_face_detection_alignment}} was used unalteredly.
The MCTNN face detector was run by the CVSSP$^d$ team.

\pphead{WVUCVL$^e$}
The WVUCVL face detection algorithm is based on detected joints on the face, inspired by CNN-based human 2D body-pose estimation methods \cite{iqbal2016multi,cao2017realtime}.
First, coordinates of the 18 main joints of the human body (e.g., shoulder center, waist, and nose) are extracted, and multi-pose estimation is applied.
Based on the five joints of the face (nose, both eyes, and both ears), frontal or side face detection is applied to the boundary of these joints, and a confidence threshold is employed.
To decrease the false accept rate, thresholds are set for checking the size of the bounding box of each face.
Finally, a skin color detector was trained on parts of the UCCS training set, and the distance between the distribution of the skin color of the bounding box from the distribution of the training set is thresholded.

\pphead{Darknet\_UOW$^f$}
The Darknet\_UOW Convolutional Neural Network (CNN) model closely follows one of the publicly available\footnote{\scriptsize\url{http://pjreddie.com/darknet}} architectures described in \cite{redmont2016yolo}, while adding a few modifications to accommodate for differences between the Visual Object Classes Challenge 2012 (VOC2012) \cite{everingham2010voc} dataset used in \cite{redmont2016yolo} and the UCCS dataset.
The Darknet\_UOW architecture consists of 22 convolutional layers followed by 2 fully connected layers, and the input size of the network has dimensions of $416 \times 416$.
Each image is divided into a $5\times5$ grid.
For each grid cell, 5 bounding boxes are predicted.
For each bounding box, the center and the dimensions are extracted, as well as a confidence represented with Intersection Over Union (IOU) between the predicted bounding box and a ground truth.

\subsection{Face Recognition}
\label{sec:participants:recognition}

\pphead{Baseline}
For the Baseline face recognition algorithm,\footref{fn:baseline} first the faces of the training set were re-detected, and face images of $64\times80$ pixels were cropped.
Histogram sequences of uniform LBP patterns \cite{ahonen04face} with $16\times16$ pixel block sizes are extracted.
A Linear Discriminant Analysis (LDA) was performed on PCA-projected features \cite{zhao1998discriminant}, using all features of unknown identities ($-1$) in one class, and each known identity in a separate class.
For enrollment of a subject (including $-1$), an average of the training set features is computed.
At test time, LBPHS features of detected faces are projected into the PCA+LDA subspace, and cosine similarities to gallery templates are computed.

\pphead{LqfNet$^c$}
A $32\times32$ pixel low-resolution CNN~\cite{herrmann2016lowresolution} is used to project each detected and downscaled face image to a discriminative 128-dimensional face descriptor.
The max-margin based network training incorporates data augmentation strategies such as blurring or adding noise to adjust the high quality training data to the low-quality domain~\cite{herrmann2016lowquality}.
About 9M face images from different public and private datasets serve as training data for the Low-quality face Network (LqfNet), while no training on the challenge data is performed.
Similar to the Baseline, for identification an LDA is learned on the gallery descriptors.
Because the LqfNet is not specifically designed to handle misdetections, the descriptor distance $d$ is weighted by the detection confidence $c$ to shape the final recognition score $s=c/d$.

\pphead{UCCS$^a$}
The UCCS contribution relies on features from the publicly available\footnote{\scriptsize\url{http://www.robots.ox.ac.uk/~vgg/software/vgg_face}} VGG Face descriptor network \cite{parkhi2015deep}, which are extracted of $224\times224$ pixel cropped images.
The enrollment is based on the Extreme Value Machine (EVM) \cite{rudd2017extreme}, which is particularly designed for open-set recognition.
Distributions of cosine distances between deep features of different identities are modeled using concepts of Extreme Value Theory (EVT), and a \emph{probability of inclusion} is computed for each enrollment feature.
Set-cover \cite{rudd2017extreme} merges several features of one identity into a single model, including a model for the unknown identities ($-1$).
We optimized EVM parameters \cite{guenther2017toward} on the validation set.
For a probe bounding box, VGG Face descriptors are computed, and the cosine similarities between probe feature and model features are multiplied with the probability of inclusion of the corresponding EVM model.

\pphead{CVSSP$^d$}
Features are extracted by two 29 layer CNNs.
The first network is trained on combined CASIA-Webface~\cite{yi2014learning} and UMD~\cite{bansal2016umdfaces} face datasets, while the other is trained on CASIA-Webface, UMD and PaSC~\cite{beveridge2013challenge} datasets.
Two feature vectors are extracted from each face and its mirror image, and merged by element-wise summation.
The template for each enrolled subject is the average of the face features extracted from the gallery images.
For the unknown subjects ($-1$), the face features are used as cohort samples for test-normalization.
During testing, the face features are extracted and the cosine similarity scores between the templates and cohort samples are computed.
For the class of unknown subjects, the similarity score is the negative of the minimum of all the templates scores.
In total, there are 1086 scores for each probe face and those scores are normalized by test-normalization.
The final score is the average of the two CNN models.

\section{Evaluation}
\label{sec:evaluation}

\begin{table*}[t!]
  \setlength{\tabcolsep}{4pt}
  \renewcommand\r[1]{\bf\textcolor{red}{#1}}
  \renewcommand\b[1]{\bf\textcolor{blue}{#1}}
  \centering\footnotesize
  \begin{tabularx}{.98\textwidth}{|r||C|C|C|C|C|c||C|C|C|C|}
    \hline
    $F\!A/FI$ & Baseline & TinyFaces & UCCS  & MTCNN & WVUCVL & Darknet\_UOW & Baseline & LqfNet & UCCS & CVSSP \\\hline\hline
       10     &   346    & \r{1008}  &   894 &     0 &    37  &   359        &   40     &   58   &\b{65}&    42 \\\hline
      100     &  4481    & \r{6466}  &  5962 &  3211 &  2664  &  2774        &  247     &  306   &\b{312}&  222 \\\hline
     1000     & 15506    &  23733    & 17350 & 18772 &\r{25124}&14721        & 2236     &  750   &\b{2892}&  2400 \\\hline
    10000     & 22802    &\r{35349}  & 30139 & 33691 & 34519  & 31441        & 3326     & 7834   & 6858 &\b{10376} \\\hline
   100000     & 26625    &\r{35789}  & 33152 & 33691 & 34519  & 34434        & 3473     & 8921   & 6858 &\b{11276} \\\hline
  \end{tabularx}
  \capt{tab:results}{Face Detection and Recognition Results}{The number of detected (left) and correctly identified (right) faces of the test set are presented for certain numbers of false accepts or false identifications, respectively. For algorithms with fewer false accepts/identifications, the total number of detected/identified faces is given. The best results are highlighted in color.}
\end{table*}

The face detection evaluation is performed on the test set where the participants provided detected face bounding boxes including confidence scores.
For face recognition, participants turned in up to ten identity predictions for each bounding box along with a similarity score for each prediction.
The evaluation scripts\footref{fn:baseline} for the validation set were given to the participants.
For all our evaluations, colors across plots correspond to identical participants.

\subsection{Face Detection}
\label{sec:evaluation:detection}
\begin{figure}[t!]
  \vspace*{-2ex}
  \centering
  \ic[.45]{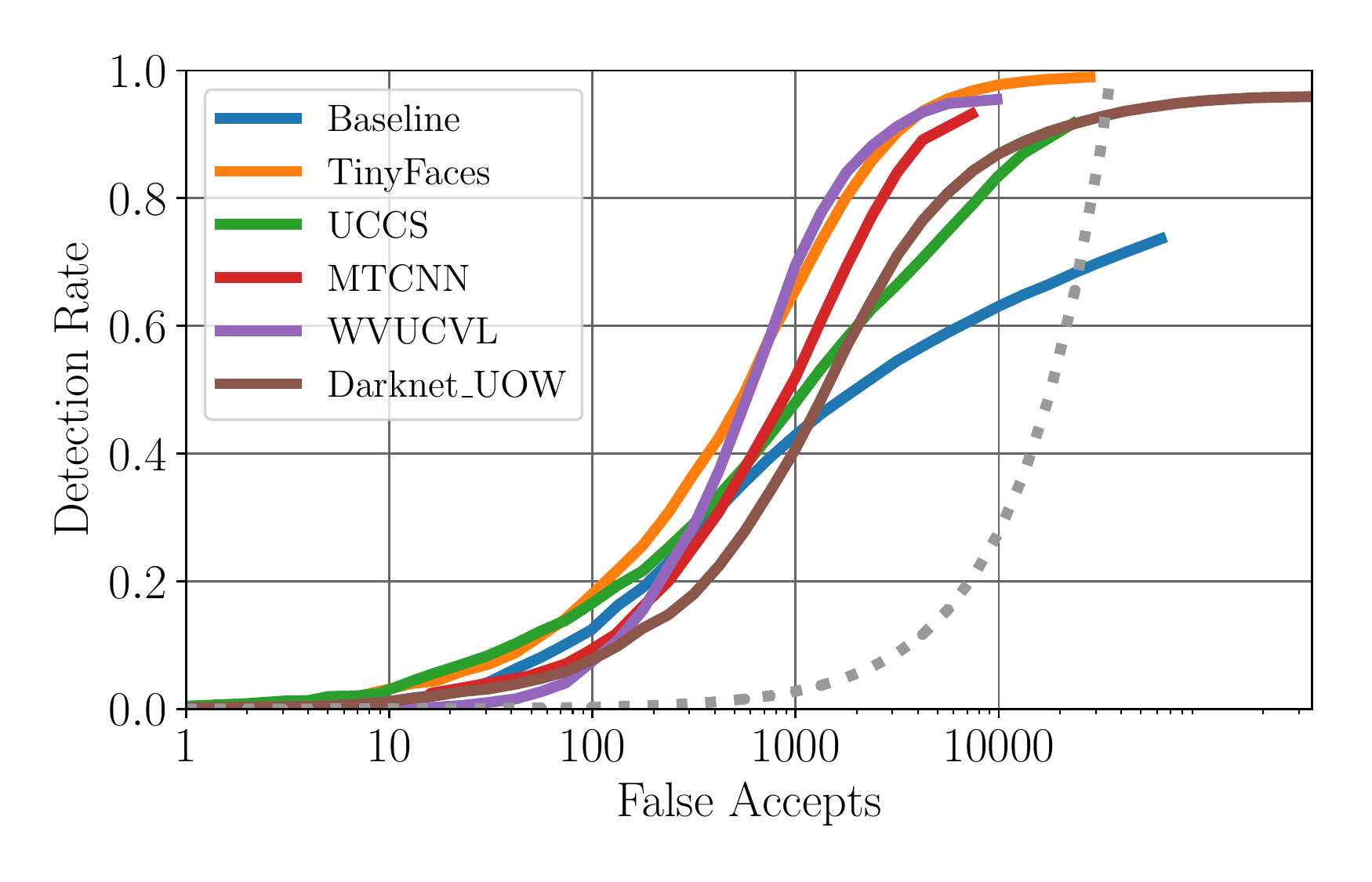}
  \capt{fig:detection}{Face Detection Evaluation}{A Free Response Operating Characteristics (FROC) curve is shown for the test set. The horizontal axis includes the number of false accepts (misdetections), while the vertical axis outlines the relative number of detected faces. The dotted gray line indicates equal numbers of correct detections and false accepts.}
\end{figure}

To evaluate the correctness of each bounding box, we use a modification of the Jaccard index, where the original Jaccard index is known as the Intersection Over Union (IOU) of the ground truth and the detected bounding box.
As the ground truth is up to four times larger than the face (cf.~\sfig{UCCS:faces}), we modify the union term to not penalize detections smaller than the ground truth:
\begin{equation}
  \label{eq:jaccard}\footnotesize
    J(G,D)  = \frac{|G \cap D|}{\max\,\bigl\{\frac{|G|}4, |G \cap D|\bigr\} + |D| - |G \cap D|} \approx \frac{|G \cap D|}{|G \cup D|}
\end{equation}
where $D$ is the area of detected bounding box and $G$ is the area of ground-truth bounding box.
Hence, when the detected bounding box $D$ covers at least a fourth of the ground-truth bounding box $G$ and is entirely contained in $G$, modified Jaccard index $J=1$ is achieved.
In our evaluation, we accept all bounding boxes with a modified Jaccard index $J \geq 0.5$.

For the face detection evaluation, bounding boxes along with their confidence scores are used in a Free Response Operator Characteristic (FROC) curve \cite{chakraborty2002statistical}.
Particularly, we split the confidence scores $c$ into positives $C^+$, i.e., where the detected bounding box overlaps with a ground truth according to \eqref{eq:jaccard}, and the negatives $C^-$ where $J(G,D) < 0.5$ for each ground-truth bounding box $G$.
For a given number of false accepts $F\!A$, we compute a confidence threshold $\theta$:
\begin{equation}
  \label{eq:detection-threshold}
  \theta = \argmax\limits_{\theta'}\ \bigl|\{c \mid c \in C^- \wedge c \geq \theta' \}\bigr| < F\!A\,.
\end{equation}
Using this threshold, the detection rate $DR$ is computed as the relative number of detected faces where the detection confidence is above threshold:
\begin{equation}
  \label{eq:detection-rate}
  DR(\theta) = \frac{\bigl|\{c \mid c \in C^+ \wedge c \geq \theta \}\bigr|}{M}\,,
\end{equation}
where $M$ is the total number of labeled faces given in \tab{Distribution}.
Finally, the FROC curve plots the $DR$ over the number of false accepts, for given values of $F\!A$.

\fig{detection} presents the results of the participants on test set, while \tab{results} contains more detailed results.
Despite the difficulty of the dataset, all face detectors (besides the Baseline) detected at least 33000 of the 36153 labeled test set faces.
Honestly, we (the challenge evaluators) were positively surprised by this result.
However, these high results can only be achieved with a relative high number of false accepts.
Still, the best performing face detectors (TinyFaces and WVUCVL) detected more than 23000 faces with 1000 false accepts, which -- given the difficulty of many of the faces -- is a very good result.
On the other hand, assuming 5 faces in each section of a $5\times5$ grid cell in the Darknet\_UOW algorithm leads to a large amount of false accepts, and might have missed some faces, i.e., when more than 5 faces were present in a certain grid cell.

\subsection{Face Recognition}
\label{sec:evaluation:recognition}
To participate in the face recognition challenge, participants were given only the raw images, i.e, without any labels.
In such an open-set scenario, a face recognition algorithm has three goals.
First -- similarly to closed-set identification -- if the probe face is of a known identity, the corresponding gallery template of that identity must have the highest similarity across all gallery templates.
Second, if the probe face is of an unknown identity, the similarities to \emph{all} gallery templates need to be small, or the probe should be labeled unknown.
Finally, when the face detector has a misdetection, that region should be handled as unknown.

\begin{figure}[t!]
  \vspace*{-2ex}
  \centering
  \ic[.45]{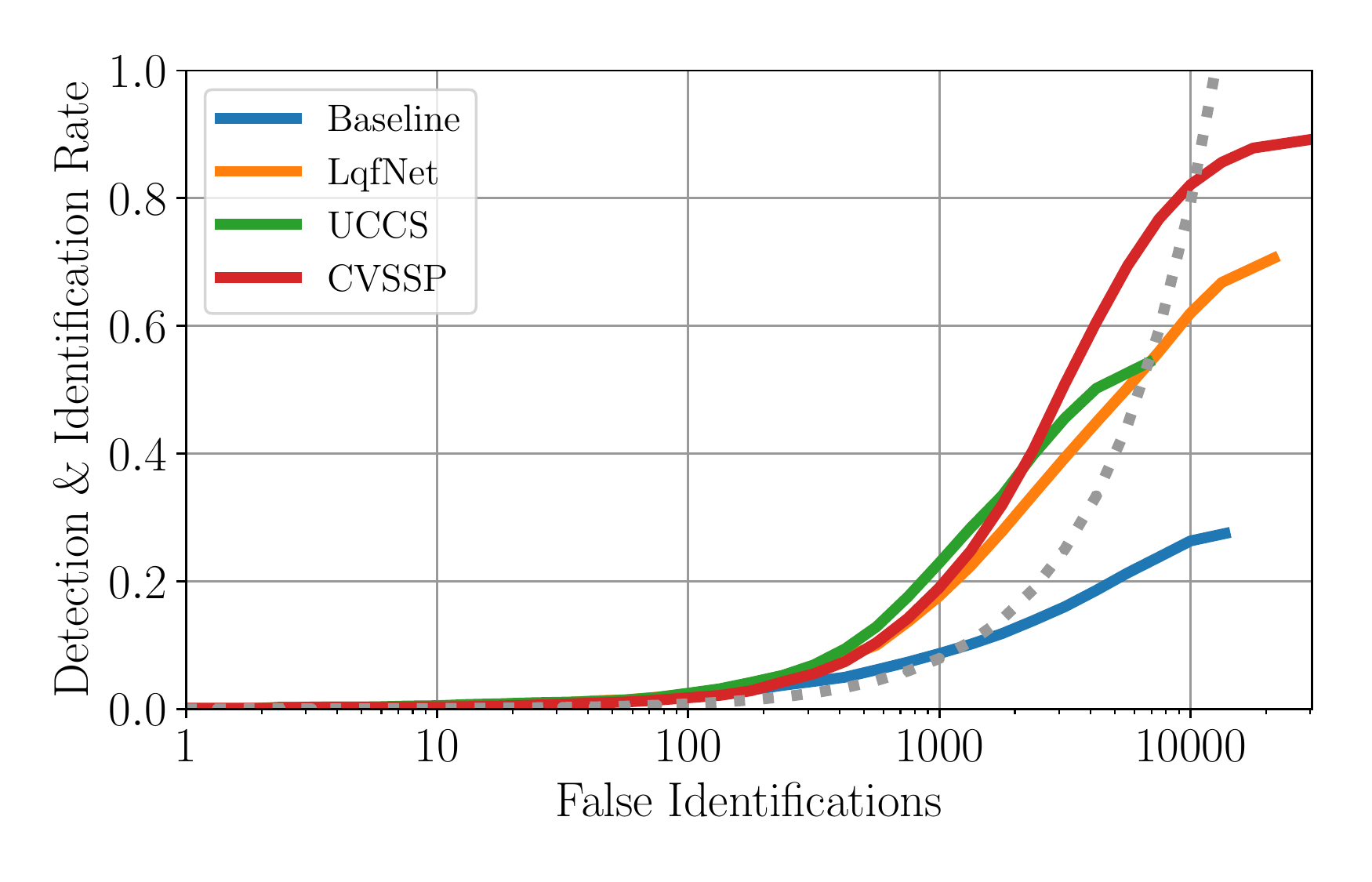}
  \capt{fig:recognition}{Face Recognition Evaluation}{A Detection and Identification Rate (DIR) curve at rank 1 is shown for the test set. The horizontal axis includes the number of false identifications, while the vertical axis outlines the relative number of correctly identified faces. The dotted gray line indicates equal numbers of correct and false identifications.}
\end{figure}

\begin{figure*}[t!]
  \vspace*{-2ex}
  \centering
  \subfloat[\label{fig:update:a}]{\ich[.104]{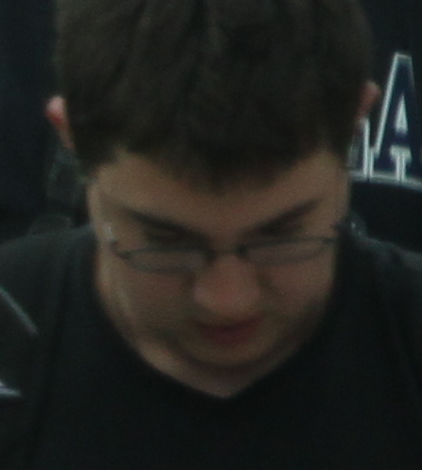}\ \ich[.104]{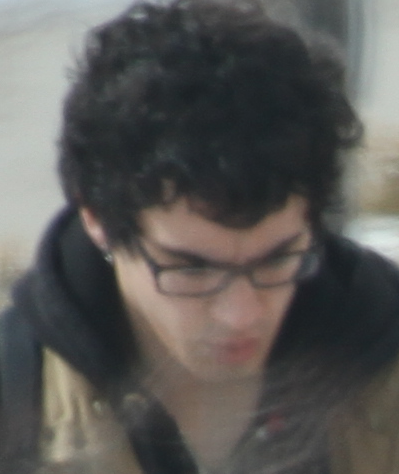}}\hspace*{1em}%
  \subfloat[\label{fig:update:b}]{\ich[.104]{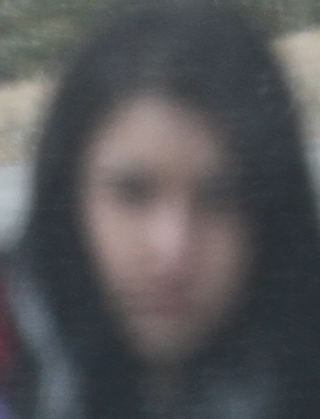}\ \ich[.104]{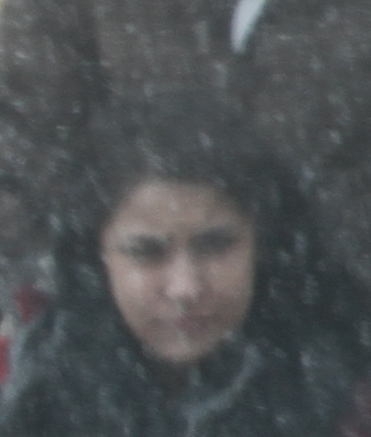}}\hspace*{1em}%
  \subfloat[\label{fig:update:c}]{\ich[.104]{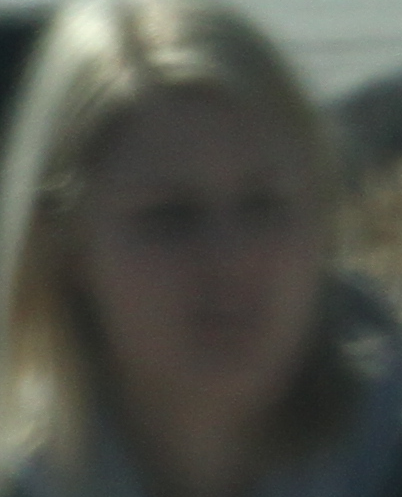}\ \ich[.104]{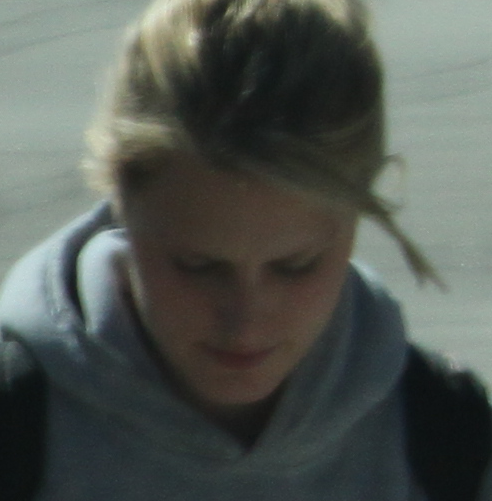}}\hspace*{1em}%
  \subfloat[\label{fig:update:d}]{\ich[.104]{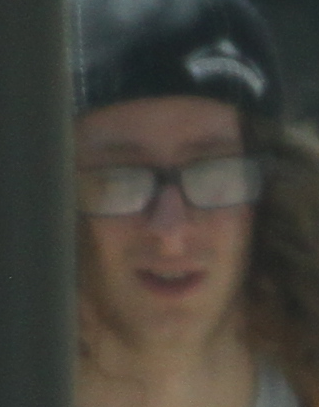}\ \ich[.104]{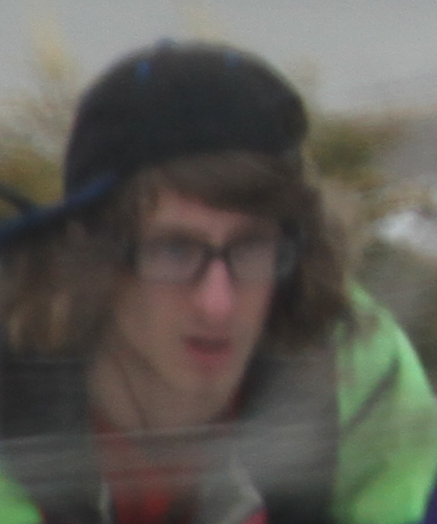}}
  \capt{fig:update}{Automatically Assigned Identities}{Some examples of automatically assigned identities are presented at the left of each pair, together with a corresponding gallery face of the newly assigned identity. New faces are assigned when all three competitors agree on the same identity. The identity in \subref*{fig:update:a} is assigned wrongly, while the remaining identities are correct. The gallery and probe images shown in \subref*{fig:update:d} were taken at different days.}
\end{figure*}

For our face recognition challenge we evaluate the participants using the Detection and Identification Rate (DIR) curve \cite{phillips2011evaluation} on rank 1.
For each probe image, we split the similarity scores $s$ of the provided bounding boxes into two groups: positives $S^+$ and negatives $S^-$.
The positive group $S^+$ contains similarity scores of correct identifications.
For each ground-truth bounding box of a known identity, the detected bounding box with the highest overlap is considered, but only if the modified Jaccard index exceeds the overlap threshold $J \geq 0.5$ defined in \sec{evaluation:detection}.
If the assigned subject label with the highest similarity score $s$ of that bounding box corresponds to the correct identity, $s$ is added to $S^+$.

The negative group $S^-$ contains similarity scores of false identifications.
It is composed of the unknown face images and the false accepts, which are labeled as a known identity.
For each ground-truth bounding box of an unknown identity, the detected bounding box with the highest overlap was considered.
If the assigned subject label with the highest similarity score $s$ of that bounding is \emph{not} $-1$, $s$ is added to $S^-$.
For false accepts, i.e., where the modified Jaccard index to every ground-truth bounding box is lower than $J<0.5$, and where the highest similarity score $s$ of that bounding box is not labeled $-1$, $s$ is appended to $\mathcal S^-$.

A decision threshold $\vartheta$ on the similarity scores can be computed for a given number of false identifications $FI$:
\begin{equation}
  \label{eq:recognition-threshold}
  \vartheta(FI) = \argmax\limits_{\vartheta'}\ \bigl|\{s \mid s \in S^- \wedge s \geq \vartheta' \}\bigl|\, < FI\,.
\end{equation}
Using this decision threshold, the relative number of correctly detected and identified persons is computed as:
\begin{equation}
  \label{eq:identification-rate}
  DIR(\vartheta) = \frac{\bigl|\{s \mid s \in S^+ \wedge s \geq \vartheta \}\bigr|}{N}\,,
\end{equation}
where $N$ is the number of known faces, cf.~\tab{Distribution}.
Finally, the DIR curve plots the detection and identification rate over the number of false identifications, for selected values of $FI$.
In our DIR plots, we do not normalize the false identifications by the total number of unknown probe faces as done by Phillips \etal{phillips2011evaluation}.
As our group of false identifications $\mathcal I^-$ includes false accepts, which are different for each face detector, normalization would favor participants with a high number of false accepts.

\fig{recognition} and \tab{results} present the results of the participants on the test set.
Given the difficulty of the dataset, from a closed-set perspective the results are impressive: almost 90\,\% of the faces were correctly identified at rank 1 by CVSSP.
From an open-set perspective, however, this comes at the price of more than 30000 false identifications.
LqfNet did not reach such a high identification rate, but still produced around 20000 false identifications.
Hence, for both algorithms up to 3 times more false identifications are raised than people are identified.
Even though the number of probe faces containing unknown identities is higher than the number of probes of known identities (cf.~\tab{Distribution}), the number of false identifications is far too high to be usable in a real scenario.
On the other hand, the UCCS algorithm, which is the best algorithm with lower numbers of false identifications, reaches only an identification accuracy of around 50\,\%.

\section{Discussion}
\label{sec:discussion}

\subsection{Database Clean-up}
\label{sec:cleanup}
The UCCS dataset is manually labeled, both the face bounding boxes and the identities.
We are aware that there are some errors in the labels, including non-marked faces in the face detection challenge, as well as faces that are labeled as $-1$, but which are actually one of the known gallery identities.
To clean up the dataset and add face bounding boxes as well as identity labels, we opted for an automatic process -- given the short time to write up this paper.
The results of this automatic process are given in parentheses in \tab{Distribution}.

To automatically mark missing faces, we use the face detectors of the participants, excluding the Baseline detector.
We select those bounding boxes that are detected by the majority of face detectors with high confidence, i.e., where detections of three algorithms overlap with IOU $\geq0.25$.
For each of the detectors, we computed a separate threshold $\theta$ at 2500 false accepts in the validation set.
To generate a new bounding box, we merged the overlapping detections, weighted with their respectively normalized detection confidence, and up-scaled them by a factor of $1.2$.
In this way, we added 1242 face bounding boxes in the validation set, and 3885 faces in the test set, which we labeled as unknown.
We have manually checked around 100 images with automatically added bounding boxes, and all of them contain valid faces.
Still, we have found that some of the faces are not marked by this automatic process.
However, for lower confidence thresholds, we found that some overlapping detections do not contain faces.

After adding these new unknown faces into the dataset, we automatically tested all faces, for which the identity was unknown.
If all three face recognition algorithms agree on the same known identity on rank 1, we assign this identity to that face.
Using this technique, 355 faces in the validation set are assigned to known identities, while in the test set it amounts to 2676 faces.
Manually checking around 100 newly assigned faces in the validation set, we found exactly one face, where the label is incorrectly assigned.
This face is shown in \sfig{update:a}, including one of the gallery faces of the assigned identity.
On the other hand, all the other inspected faces are correctly labeled, including images from different days, see \sfig{update:d} for an example.
Note that we could not update the masked identities, cf.~\tab{Distribution}.

\begin{figure*}[t!]
  \vspace*{-2ex}
  \centering
  \subfloat[\label{fig:updated:detection}Face Detection]{\ic[.45]{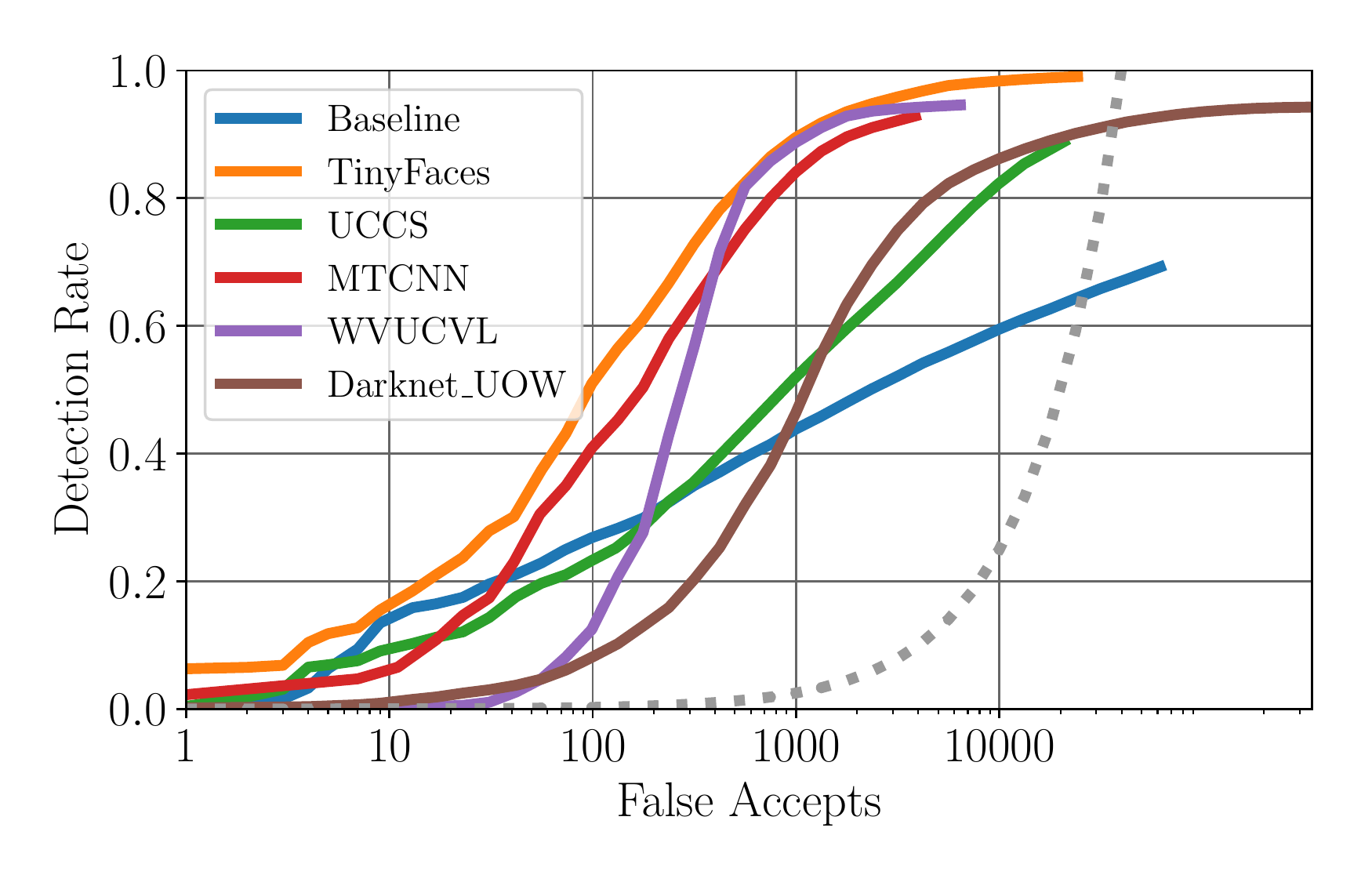}}\qquad%
  \subfloat[\label{fig:updated:recognition}Face Recognition]{\ic[.45]{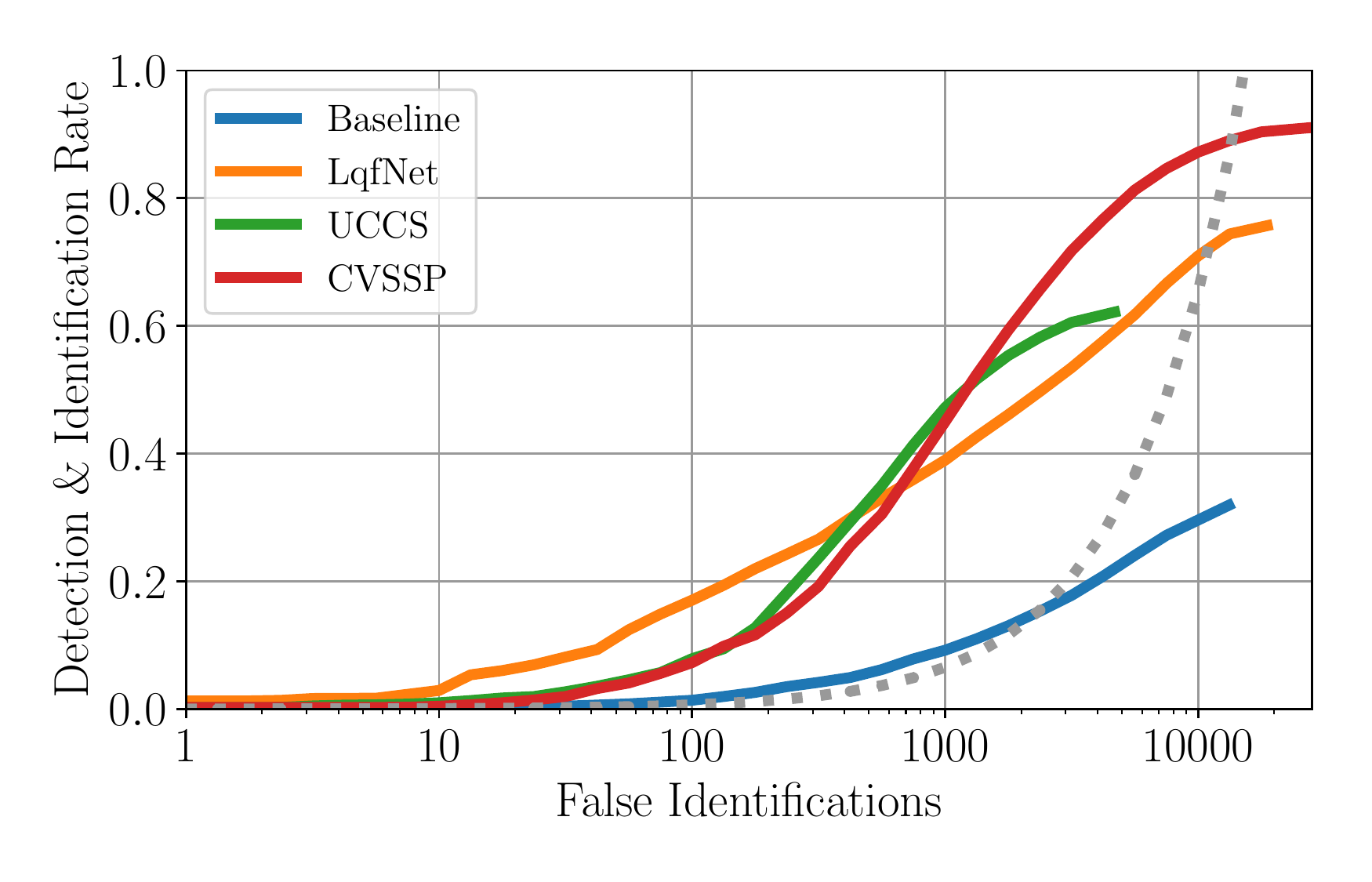}}
  \capt{fig:updated}{Updated Evaluation}{FROC and DIR curves of all participants are displayed on the test set for the automatically updated ground-truth labels.
  The dotted gray line indicates equal numbers of correct detections/identifications and false accepts/identifications.}
\end{figure*}

\begin{figure*}[t!]
  \vspace*{-2ex}
  \centering
  \subfloat[\label{fig:day:same}Same Day]{\icp[.45]{1}{DIR-SD-test}}\qquad%
  \subfloat[\label{fig:day:different}Different Day]{\icp[.45]{2}{DIR-SD-test}}
  \capt{fig:day}{Same vs. Different Day}{DIR curves are comparing templates and probes taken on \subref*{fig:day:same} the same and \subref*{fig:day:different} different days. The horizontal axis includes the number of false identifications, while the vertical axis outlines the rate of correctly identified faces.}
\end{figure*}

When evaluating the participants' detection and recognition algorithms on the updated labels of the test set, we can see that the results generally improve.
In the FROC curve in \sfig{updated:detection}, out of the 40038 faces that are labeled in the updated test set, two algorithms can detect around 90\,\% of them at 1000 false accepts.
Interestingly, in comparison to \fig{detection}, MTCNN is almost able to catch up with TinyFaces and WVUCVL.
Finally, TinyFaces detects almost all faces, but on the cost of more than 30000 false accepts.

The DIR curve in \sfig{updated:recognition} shows a difference to \fig{recognition} with respect to the order of the participants on lower numbers of false identifications, where now the LqfNet algorithm performs considerably better.
Generally, we can see an improvement of all algorithms, and at 1000 false identifications now both UCCS and CVSSP are able to identify around half of the known subjects correctly.

\subsection{Analysis of Time Differences}
\label{sec:time-differences}
Though the UCCS dataset was collected over more than two years, many people appear only in a single day.
The training and validation sets are built such that for most of the known identities all faces stem from a single day.
In the test set, we have put some images of known subjects, which are taken at a different day than present in the training set.
The updated protocol of the test set contains 20647 faces of 932 identities taken at the same day as the training set faces of the corresponding identities, and 14432 faces of 209 identities with different days.

In \fig{day} we show the difference of the participants' results between the two sets of probe faces in the updated test set.
As we could not split the unknown faces into same or different days, the same false identifications are used in both plots.
It comes as no surprise that the different day faces are harder to be identified, the rates in \sfig{day:different} are considerably lower than in \sfig{day:same}.
At 1000 false identifications, all algorithms are below 20\,\% DIR, and below 3\,\% at 100 false identifications.
While the Baseline is practically useless when images are taken at different days, all the participating algorithms are not much more reliable either.

\subsection{Analysis of False Identifications}
\label{sec:false-alarms}
\begin{figure*}[t!]
  \vspace*{-2ex}
  \centering
  \subfloat[\label{fig:false-alarms:in-train}Masked in Training]{\icp[.32]{1}{KUU-test}}\ %
  \subfloat[\label{fig:false-alarms:not-in-train}Masked not in Training]{\icp[.32]{2}{KUU-test}}\ %
  \subfloat[\label{fig:false-alarms:misdetections}False Accepts]{\icp[.32]{3}{KUU-test}}
  \capt{fig:false-alarms}{Evaluation of False Identifications}{In \subref*{fig:false-alarms:in-train} and \subref*{fig:false-alarms:not-in-train} the percentage of correctly rejected masked identities are plotted over the number of correctly identified faces. In \subref*{fig:false-alarms:misdetections} the relative number of correctly rejected false accepts (misdetections) is plotted over the number of correctly identified faces, normalized by the total number of false accepts per detection algorithm.}
\end{figure*}

In the evaluations above, false identifications are computed jointly from unknown faces and false accepts (misdetections).
To evaluate the impact of each of these on the performance, we split the false identifications.
Here, we only use the \emph{masked} faces, i.e., which have been given as $-1$ to the participants, but where we know the identity labels.
The masked faces are further split up into masked identities that are in the training set, and masked identities that are not, cf.~\tab{Distribution}.
The evaluation is performed on the automatically cleaned dataset.

To have a better comparable evaluation, we plot correctly rejected masked faces or false accepts over the correctly identified known identities.
Hence, the similarity score threshold $\vartheta$ is now computed over $\mathcal S^+$ (cf.~\sec{evaluation:recognition}), while the correct rejection rate is computed as:
\begin{equation}
  \label{eq:rejection-rate}
  CRR(\vartheta) = \frac{\bigl|\{s \mid s \in \mathcal S^= \wedge s < \vartheta \}\bigr|}{\bigl|\{\mathcal S^= \}\bigr|}
\end{equation}
where $\mathcal S^=\!\subset\!S^-$ are the corresponding false identifications.

In \sfig{false-alarms:in-train} we plot the number of correctly rejected masked images where the identities are included in the training set, while \sfig{false-alarms:not-in-train} contains the masked images where subjects are not part of the training set.
As algorithms that model unknown faces as a separate class during training, the LqfNet and UCCS algorithms decrease rejection capabilities for unknown identities seen during training (\sfig{false-alarms:not-in-train}), which is counterintuitive, but we attribute this behavior to some wrong labels in the training set.
On the other hand, the CVSSP algorithm did not make use of the training set and, thus, its performance is stable with respect to the masked identities.

Finally, \sfig{false-alarms:misdetections} shows how the algorithms deal with their respective false accepts.
There, the UCCS and LqfNet algorithms are able to reject almost all false accepts, even with low threshold $\vartheta$ (high number of correct identifications), while CVSSP starts dropping the correct rejection rate and finally assigns a known identity label to each of its false accepts at low thresholds.

\section{Conclusion}
\label{sec:conclusion}
We have evaluated the participants' results of the unconstrained face detection and the open-set face recognition challenge.
We were surprised by the quality of the face detectors, and the closed-set recognition capabilities of the algorithms on our very difficult dataset.
However, open-set face recognition, i.e., when face recognition algorithms are confronted with unknown faces and misdetections is far from being solved, especially when probe faces are taken at different days than the gallery faces.

For this paper, we automatically updated our ground-truth labels by majority voting of the participants' algorithms.
With this, we surely have missed some of the faces, and some identity labels are definitely wrong.
We will use the participants' results to start a semi-manual re-labeling of the data, i.e., we propose overlapping bounding boxes to a human observer who decides whether a face is seen, or whether two face images show the same identity.
The training and validation sets will be made public after this process is finished, while the test set will be kept secret and used in further challenges.

For the present competition, we provided a biased evaluation protocol, i.e., the training set is identical to the enrollment set.
As we have seen, already with this biased protocol open-set face recognition is difficult.
More unbiased evaluations would split off several identities into a training set, and enrollment and probing would be performed on a different set of identities.
We will investigate on such an unbiased protocol in future work.

\section*{Acknowledgment}
This research is based upon work funded in part by NSF IIS-1320956 and in part by the Office of the Director of National Intelligence (ODNI), Intelligence Advanced Research Projects Activity (IARPA), via IARPA R\&D Contract No. 2014-14071600012. The views and conclusions contained herein are those of the authors and should not be interpreted as necessarily representing the official policies or endorsements, either expressed or implied, of the ODNI, IARPA, or the U.S. Government. The U.S. Government is authorized to reproduce and distribute reprints for Governmental purposes notwithstanding any copyright annotation thereon.

{\small
\bibliographystyle{ieee}
\bibliography{challenge}
}

\end{document}